%% file: main.tex
\documentclass[runningheads]{llncs}

\usepackage{graphicx}
\usepackage{amsmath}
\usepackage{amssymb}
\usepackage{booktabs}
\usepackage{sidecap}
\usepackage{enumitem}
\setlist[itemize]{parsep=0pt,topsep=2pt,itemsep=2pt}
\setlist[enumerate]{parsep=0pt,topsep=1pt,itemsep=1pt}
\usepackage{adjustbox}
\usepackage[dvipsnames]{xcolor}

\usepackage{adjustbox}

\definecolor{cvprblue}{rgb}{0.21,0.49,0.74}

\usepackage[breaklinks,colorlinks,citecolor=cvprblue]{hyperref}

\usepackage[capitalize]{cleveref}
\usepackage{standalone}

\crefname{section}{Sec.}{Secs.}
\Crefname{section}{Section}{Sections}
\Crefname{table}{Table}{Tables}
\crefname{table}{Tab.}{Tabs.}

\input{commands}

\title{\titletext}
\titlerunning{Multi-View Projection for UDA in 3D Semantic Segmentation}

\author{Andrew Caunes\inst{1,2} \and
Thierry Chateau\inst{1} \and
Vincent Frémont\inst{2}}

\authorrunning{A. Caunes et al.}

\institute{Logiroad, Nantes, France \\
\and
LS2N - Ecole Centrale de Nantes, France\\
}

\begin{document}
\maketitle
\input{sections/abstract}

\input{sections/introduction}

\input{sections/related}

\input{sections/method}

\input{sections/experiments}

\input{sections/conclusion}

{
    \small
    \bibliographystyle{splncs04}
    \bibliography{bib}
}

\end{document}

%% file: commands.tex
\usepackage{graphicx}
\usepackage{amsmath}
\usepackage{amssymb}

\usepackage[utf8]{inputenc} 
\usepackage[T1]{fontenc}    
\usepackage{lmodern}        
\usepackage{url}            
\usepackage{booktabs}       
\usepackage{amsfonts}       
\usepackage{nicefrac}       
\usepackage{microtype}      
\usepackage[dvipsnames]{xcolor}         

\usepackage{amssymb}
\usepackage{pifont}
\usepackage{fixmetodonotes}
\usepackage{colortbl}
\usepackage{rotating}
\usepackage{scalerel}

\usepackage{multirow}
\usepackage[normalem]{ulem}
\usepackage{cancel}

\usepackage{bm}
\usepackage{adjustbox}
\usepackage{array}
\newcolumntype{R}[2]{
    >{\adjustbox{angle=#1,lap=1.3\width-(#2)}\bgroup}
    l
    <{\egroup}
}

\usepackage[linesnumbered,ruled,vlined]{algorithm2e}
\usepackage{algpseudocode}

\makeatletter 
\@namedef{ver@everyshi.sty}{}
\makeatother
\usepackage{tikz}
\usetikzlibrary{positioning,shapes}

\definecolor{MyGreen}{RGB}{0, 180, 0}
\definecolor{MyRed}{RGB}{180, 0, 0}
\definecolor{MyBlue}{RGB}{30, 0, 180}
\definecolor{MyGrey}{RGB}{82.75, 82.75, 82.75}

\definecolor{skcar}{RGB}{100,150,245}
\definecolor{skbicycle}{RGB}{255, 200, 0}
\definecolor{skmotorcycle}{RGB}{255, 120, 0}
\definecolor{sktruck}{RGB}{80, 30, 180}
\definecolor{skotherv}{RGB}{0, 0, 255}
\definecolor{skpedestrian}{RGB}{255,30,30}
\definecolor{skdrivable}{RGB}{255, 0,255}
\definecolor{sksidewalk}{RGB}{75, 0,75}
\definecolor{skterrain}{RGB}{150, 240, 80}
\definecolor{skvegetation}{RGB}{0, 175, 0}
\definecolor{skbuilding}{RGB}{255, 200, 0}
\definecolor{mygreen}{RGB}{0, 170, 0}

\definecolor{tsne_source}{rgb}{0.86, 0.3712, 0.33999999999999997}
\definecolor{tsne_target}{rgb}{0.33999999999999997, 0.8287999999999999, 0.86}

\newcommand{\cmark}{{\textcolor{MyGreen}{\ding{51}}}}

\newcommand{\method}{{SEG3DBYPC2D}}
\def \method{Seg3DbyPC2D \xspace}

\newcommand{\synth}{{SL}}

\newcommand{\sk}{{SK}} 
\newcommand{\ns}{{NS}} 
\newcommand{\skns}{{SK$_{10}$}} 
\newcommand{\sksyn}{{SK$_{19}$}}

\newcommand{\synthtosk}{{\DAsetting{\synth}{\sksyn}}}

 \newcommand{\titletext}{Multi-View Projection for Unsupervised Domain Adaptation in 3D Semantic Segmentation}

\newcommand{\second}[1]{\cellcolor{orange!25}{#1}}
\newcommand{\best}[1]{\cellcolor{red!25}{#1}}

\usepackage{stmaryrd}
\usepackage{trimclip}

\makeatletter
\DeclareRobustCommand{\shortto}{
  \mathrel{\mathpalette\short@to\relax}
}

\newcommand{\short@to}[2]{
  \mkern2mu
  \clipbox{{.5\width} 0 0 0}{$\m@th#1\vphantom{+}{\shortrightarrow}$}
  }
\makeatother

\newcommand{\DAsetting}[2]{{#1}$\rightarrow${#2}}

%% file: sections/abstract.tex
\begin{abstract}
    3D semantic segmentation plays a pivotal role in autonomous driving and
    road infrastructure analysis, yet state-of-the-art 3D models are prone
    to severe domain shift when deployed across different datasets. 
    In this paper, we
    propose an Unsupervised Domain Adaptation approach where a
    3D segmentation model is trained on the target dataset 
    using pseudo-labels generated by a novel multi-view projection 
    framework.
    Our approach first aligns Lidar scans into coherent 3D scenes and
    renders them from multiple virtual camera poses to create 
    large-scale synthetic 2D semantic segmentation datasets in various modalities.
    The generated datasets are used to train an 
    ensemble of 2D segmentation models in 
    point cloud view domain on each modality.
    During inference, the models process 
    a large amount of views per scene; the resulting logits are 
    back-projected to 3D with a depth-aware voting scheme 
    to generate final point-wise labels. These labels are then used 
    to fine-tune a 3D segmentation model in the target domain.
    We evaluate our approach Real-to-Real on the nuScenes and SemanticKITTI
    datasets.
    We also evaluate it Simulation-to-Real with the SynLidar
    dataset.
    Our contributions are a novel method that achieves state-of-the-art results 
    in Real-to-Real Unsupervised Domain Adaptation, and we also demonstrate an 
    application of our method to segment `rare classes',
    for which target 3D annotations are not available, by only using 2D annotations 
    for those classes and leveraging 3D annotations for other classes in a source domain.

    \keywords{3D Semantic Segmentation \and Unsupervised Domain Adaptation \and Autonomous Driving \and Lidar \and Multi-view Projection}
\end{abstract}

%% file: sections/introduction.tex
\section{Introduction}

\input{figures/overview_method.tex}

3D semantic segmentation of Lidar scenes is 
crucial for applications such as autonomous
driving (AD) and road infrastructure analysis.
In particular, many applications require
models to be able to generalize to unseen
data, which may not be from the same 
domain as the training data, e.g.
if the data was acquired in a different location
or with a different sensor.
The main approach to mitigate this problem
is Unsupervised Domain Adaptation (UDA).

Most existing methods are tailored for AD,
thus 
making concessions on accuracy to
guarantee real-time performance on 
single or few Lidar scans \cite{shaban_lidar-uda_2023,saltori_compositional_2023,michele_saluda_2023,zhao_unimix_2024}.

We propose a method that leverages aligned scenes
to generate high quality pseudo-labels for the 
target domain, which are used to fine-tune a final 3D
segmentation model pre-trained on the source domain.
Our approach reaches state-of-the-art performance
in UDA while introducing no computational overhead
to the final 3D segmentation model.
Our pseudo-label generation process  
belongs to the multi-view projection 
family of methods, where 3D scenes are analyzed via multiple 
2D representations. We position virtual
cameras around 
aligned Lidar scenes to generate
rendered views (virtual camera images) from 
various perspectives. While similar methods 
use real camera images for training and 
inference \cite{genova_learning_2021},
we create large-scale synthetic 2D semantic segmentation
datasets by rendering both 
the 3D scenes and their corresponding
ground truth annotations. 
This eliminates the need for additional camera sensors
and allows to freely choose the optimal parameters for 
selecting and rendering the views.

We demonstrate the effectiveness of our method
for UDA in Real-to-Real tasks
on the nuScenes (NS) \cite{caesar_nuscenes_2020} and
SemanticKITTI (SK) \cite{behley_semantickitti_2019} datasets 
(\DAsetting{\ns}{\sk} and \DAsetting{\sk}{\ns}),
and in a Simulation-to-Real task on the SynLidar (SL) \cite{xiao_transfer_2021} dataset
(\DAsetting{\synth}{\sk}).
We also demonstrate a potential application of our method to segment `rare classes',
where 3D annotations are not available for the desired target classes, but 2D annotations 
are available along with 3D annotations for other classes.
Finally, we conduct a full ablation study to
verify the importance of each component of our method.
\textbf{Our main contributions are the following:}
\begin{itemize}
    \item A method for Unsupervised Domain Adaptation in 3D semantic segmentation reaching state-of-the-art performance in 
    
    nuScenes $\rightarrow$ SemanticKITTI and SemanticKITTI $\rightarrow$ nuScenes
    \item A novel multi-view projection framework 
    for pseudo-label generation, with dataset generation capabilities 
    \item A demonstration of an application of our method to segment `rare classes' 
    for which 3D annotations are not available
\end{itemize}

%% file: figures/overview_method.tex
\begin{figure}[th!]
\centering
\includegraphics[width=0.9\textwidth]{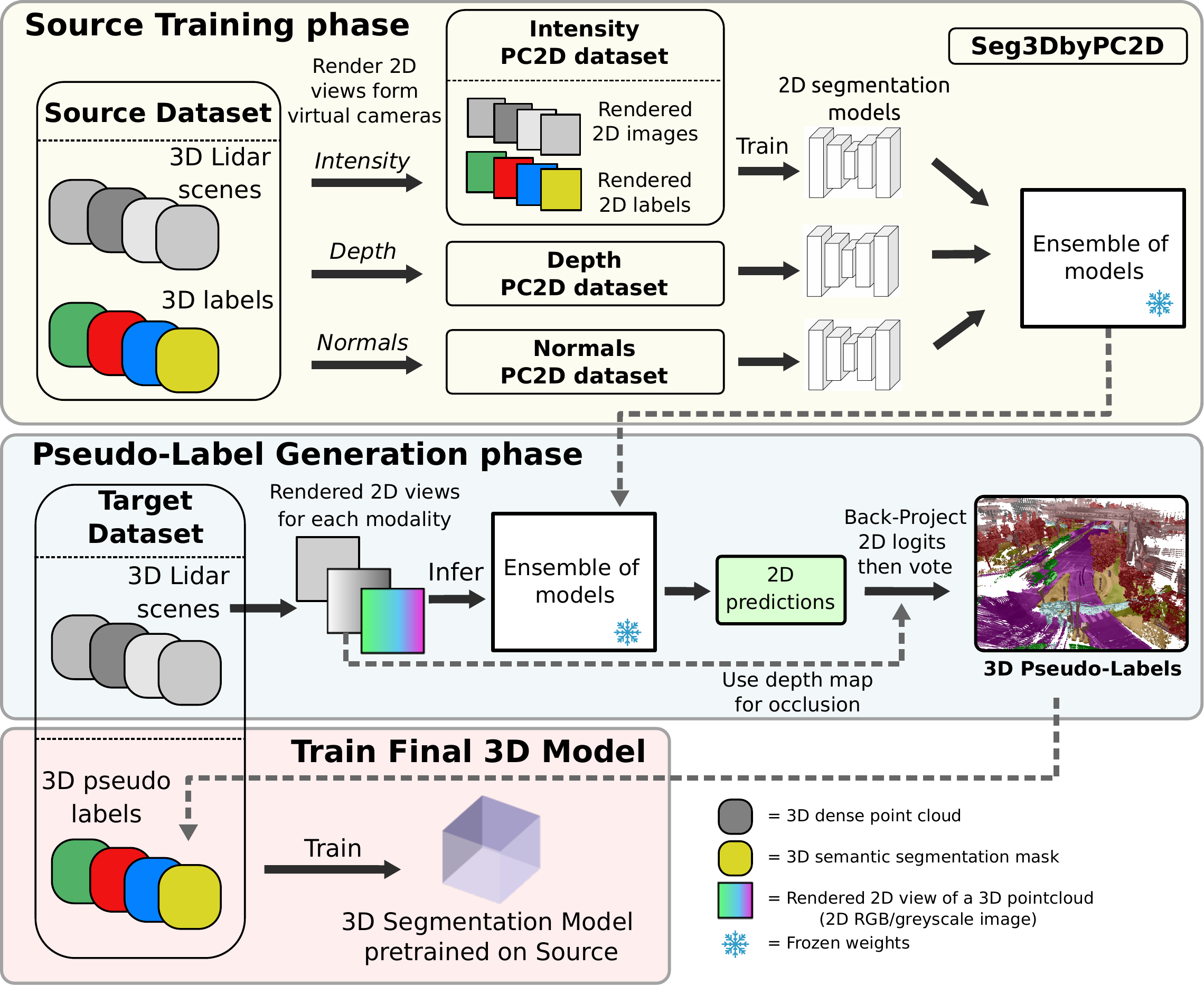}
\caption{\textbf{Overview of {\method}.} 
Unsupervised Domain Adaptation for 3D semantic segmentation
is achieved by fine-tuning a 3D model pretrained
on the source dataset on pseudo-labels 
generated for the target domain.
To generate these pseudo-labels, 
a multi-view projection method is used.
\textit{Source Training Phase (yellow).}
First, starting from 3D scenes (LiDAR scans accumulated in a common frame) 
and 3D segmentation masks, virtual camera poses
are sampled around each scene. 2D images along with 2D segmentation masks 
are then rendered from the poses in a chosen modality (e.g. Lidar intensity colors). 
See Fig.~\ref{fig:PC2D_generation} for details on this dataset generation step. 
Each generated dataset is then used to train an individual 2D semantic 
segmentation model, which together form an ensemble.
\textit{Pseudo-label generation phase (blue).} 
For each target 3D scene, pseudo-labels are generated 
by following the same view rendering process as in the training phase
and running them through the ensemble.
The resulting 2D logits are back-projected to 3D and accumulated as votes.
The final 3D mask is obtained by assigning the most voted class to each point.
\textit{Target Training Phase (red).} Finally, a 3D model pretrained on
the source dataset is fine-tuned on the pseudo-labels.
}
\label{fig:overview}
\vspace*{-1mm}
\end{figure}

%% file: sections/related.tex
\section{Related works}

\subsection{3D semantic segmentation}

3D semantic segmentation is the task of assigning a 
semantic label to each point in a 
3D point cloud. It is generally done either on Lidar point clouds 
\cite{caesar_nuscenes_2020,behley_semantickitti_2019}, 
or RGB-D images \cite{dai_scannet_2017}. 
Methods are usually divided in 3D-based and 2D-based. 
3D models directly operate on the point cloud
using 3D sparse convolutions 
\cite{choy_4d_2019-2}, 
while 2D models re-use 
2D image segmentation methods by projecting 
the point cloud to a 2D plane 
\cite{lang_pointpillars_2018}. 
Importantly, LiDAR-based 3D semantic segmentation models are particularly 
sensitive to domain shift due to sensor-specific sampling and sparsity effects, 
often more so than 2D vision models \cite{shaban_lidar-uda_2023,lambert_mseg_nodate}.

 2D approaches generally
project the point cloud to a single 2D image with channels 
representing various features, e.g. height, intensity, density, etc.
Various projection approaches exist, including 
Bird's Eye View (BEV) \cite{lang_pointpillars_2018} 
and Range View (RV) \cite{wu_squeezesegv2_2019}.
Multi-view projection methods try to leverage 2D
models even further by repeatedly using the same model on multiple
views of the 3D scene.

\subsection{Multi-view projection for segmentation}
\label{related:multi}
The goal of multi-view projection methods is to extract as much
knowledge from a 2D model as possible for 3D segmentation. This 
is done by inferring on multiple views of the 3D scene
and then merging the resulting 2D segmentations.
The problem then decomposes itself into two parts:
\begin{enumerate}[label=P\arabic*.]
    \item Which projections to use for the views and with which inference model?
    \item How to merge the 2D segmentation masks to get unique 3D labels?
\end{enumerate}

Notice that the second problem is equivalent to the 2D-3D label propagation problem
\cite{wang_label_nodate}. It is generally solved using either 
a voting scheme \cite{vedaldi_virtual_2020,mccormac_semanticfusion_2016,caunes_3d_2024}, where 
either the 2D masks or the probabilities/logits of the model are accumulated
on points as votes, or with more complex methods such 
as Neural Networks \cite{robert_learning_2022,ma_multi-view_2017,wang_ldls_2019}.

Our proposed method uses a fast voting scheme on accumulated logits 
from a single 2D model. 

In this paper, we focus on problem P1.  
Initially, \cite{su_multi-view_2015} proposed
multi-view projection for indoor RGB-D scenes, where the views 
 are simply 
the RGB-D images themselves.
For outdoor Lidar scenes, the same reasoning led \cite{genova_learning_2021} to use
RGB camera images as views, with cameras calibrated to the Lidar \cite{neuhold_mapillary_2017}. 
However, this approach is 
limited as it requires using
 additional sensors, and the performance is 
tied to the quality of the sensor 
calibration.
Mitigating this, \cite{caunes_3d_2024} proposed to infer on rendered views 
of 3D scenes 
with an out-of-domain 2D model trained on camera images. 
This has the advantage
of not requiring 3D labels or additional sensors, but comes with several limitations:
\begin{itemize}
    \item The domain gap between camera images and views of Lidar
     point clouds is significant, which negatively 
     impacts 2D model performance.
    \item The multi-view capability of the method is limited, 
    as virtual camera placement must remain close to 
    physically plausible viewpoints, 
    e.g., those of real road users.
    \item Because aligning Lidar scans is essential 
    for obtaining dense 3D scenes,
    the method is limited
    to static classes, as dynamic classes often appear 
    with a so-called `\textit{shadow effect}' in the aligned scene,
    which is out of the camera images' domain.
\end{itemize}

We propose to solve these three limitations by leveraging 3D annotations. 
This makes our method more dependent on 3D annotated data, but allows 
it to achieve significantly better performance and handle dynamic classes.
Closest to our approach, \cite{vedaldi_virtual_2020} proposes to use 
3D annotations to generate 2D synthetic datasets to train a 2D model 
for downstream multi-view projection for indoor RGB-D scenes. 
Our method distinguishes itself in
its application to outdoor Lidar scenes, where much additional processing is 
required to obtain satisfying views, e.g. scan aligning,
as well as in the choices for the feature rendering and view selection.
We propose an ensembling mechanism for handling the modality choice.
We also demonstrate the potential of such methods for pseudo-label generation
for unsupervised domain adaptation.

\subsection{Unsupervised Domain Adaptation}
UDA aims to adapt a model trained on a labeled source domain 
$(X_S, Y_S)$ to perform well on an unlabeled target domain $X_T$, where no ground-truth labels are available. 
Unlike supervised adaptation, only the source labels and unlabeled target samples $X_{T\_train}$ 
can be exploited, while the evaluation is performed on a disjoint set $X_{T\_test}$. 
This setting is particularly relevant for autonomous driving, where annotated data 
is costly to acquire, but raw scans from the deployment domain can be collected at scale.  

We compare our approach to state-of-the-art UDA methods for 3D semantic segmentation, which mostly 
rely on self-training, temporal consistency, or compositional data augmentation.  
T-UDA \cite{gebrehiwot_t-uda_2023} leverages temporal coherence by enforcing 
sequential point cloud consistency and cross-sensor geometric alignment within a mean-teacher framework.  
Lidar-UDA \cite{shaban_lidar-uda_2023} simulates varying scan patterns via random beam dropping 
and improves pseudo-label reliability through cross-frame consistency and self-training.  
CoSMix \cite{saltori_compositional_2023} introduces compositional mixing across domains, 
generating hybrid point clouds to reduce the distribution gap between source and target.  
SALUDA \cite{michele_saluda_2023} further improves self-training by enforcing domain-invariant 
geometric constraints on pseudo-labels.  
Finally, UniMix \cite{zhao_unimix_2024} extends mixing-based approaches with a unified framework 
that interpolates both features and labels across domains, enhancing robustness to domain shift.  

Our method differs by relying entirely on a multi-view 2D projection pipeline to generate 
pseudo-labels for the target domain. These pseudo-labels are used to fine-tune a 
3D segmentation model pretrained on the source domain. 
This approach reaches state-of-the-art
performance in UDA.

%% file: sections/method.tex
\section{Method}
\label{method}
\input{figures/PC2D_generation.tex}

An overview of our method, which we call \method{}, is shown in Fig.~\ref{fig:overview}.
To achieve UDA in 3D semantic segmentation, we simply
pretrain a 3D model on the source dataset, then fine-tune
it on the target dataset using pseudo-labels.
Our main contribution is therefore our pseudo-label generation pipeline,
which we describe in detail in this section.

In the following, we use the terms `\textit{scene}' to refer to a point cloud
resulting from the accumulation of Lidar scans into a common frame, 
and `\textit{view}' to refer to a 2D image rendered from a 3D scene 
using a virtual camera. 

\subsection{Pseudo-label Generation}

Our method can be decomposed in the source training phase and the target 
inference phase. In both phases, we obtain 3D scenes and sample virtual camera 
poses to render views of the scenes.
In the source training phase, we use these rendered views, along
with rendered views of the 3D segmentation masks, to generate synthetic
segmentation datasets (PC2D) that will be used to train an ensemble 
of 2D models.
In the target inference phase, the views are used for inference with
the previously trained ensemble of models, yielding pseudo-labels
for the target domain.
The choices of parameters for Virtual Camera Placement (VCP) and 
view rendering from Multiple Modalities (MMods) are crucial and are 
detailed in the following
subsections.
To obtain final pseudo-labels in the target domain, our
multi-view projection approach is based on \cite{caunes_3d_2024},
which we improve with occlusion handling.

\subsubsection{Training with 3D annotations in 2D (PC2D)}

\label{method:PC2D}

A major contribution to our work compared to existing
multi-view projection methods is the training of 2D models
in \textit{rendered view} domain. 
To that end, we generate large-scale synthetic datasets
of 2D views of 3D scenes along with 2D segmentation
masks, as shown in Fig.~\ref{fig:PC2D_generation}.
We call these PC2D datasets.

Virtual camera poses are sampled arbitrarily around the 3D scene, 
which are then used 
to render 2D views.
This way, we can generate an arbitrary number of (image, label)
pairs, yielding large-scale datasets. 
Training on such datasets can be seen as a powerful 
data augmentation technique. 
The main limitation to this is that a limited number 
of annotated scenes will yield samples with less diversity 
as the total number of training samples increases.

The use of PC2D datasets allows to train the models in 
a \textit{rendered view} source domain, very close to the target
inference domain, yielding greater freedom over the 
view generation steps compared to existing methods.
For example, generating bird's eye view depth maps is possible, 
while views in \cite{caunes_3d_2024},
are limited to `road user viewpoint' rendering of Lidar intensity.
The two following 
subsections describe how we choose
the virtual camera placement and the rendering modalities
to obtain the best performance.

\subsubsection{Virtual Camera Placement}
\label{method:VCP}

We propose to optimize the view generation by choosing relevant virtual camera poses.

After experimentation, we retain 4 types of camera poses based on the 
original Lidar sensor poses, which are already describing 
the trajectory of the ego vehicle
along the scene. An illustration of the
4 types of camera pose is shown in Fig.~\ref{fig:PC2D_generation}, in the VCP module, and we describe them below:
\begin{itemize}
    
    \item \textbf{carView}: The camera is placed at a road user's perspective, i.e. at the approximate
    height of a car (random yaw can be applied).
    \item \textbf{conicView}: The camera is placed on the basis of a large imaginary cone pointing downwards,
    with the apex at the Lidar sensor position. The camera points to the apex of the cone. This 
    corresponds to a `drone' or `lamp post' view of the scene, and is also the closest to the views
    that human annotators would typically use to annotate a scene.
    \item \textbf{topView}: The camera is placed directly above the Lidar position, looking downwards,
    with variable height and roll. This corresponds to the usual `bird's eye view' (BEV) of the scene.
    \item \textbf{bottomView}: The camera is placed under the scene, looking upwards.
    This is the opposite of a BEV.
    Rendering Lidar point clouds from 
    below allows to clearly see the colors of the flat surfaces of the ground without
    occlusions from the objects above.
\end{itemize}

\subsubsection{Multi-Modalities}
\label{method:MMods}

Similar to camera pose generation, PC2D datasets allow 
us to choose the visualization modality
for the training and inference images, i.e. the rendered colors. 
While multi-view projections methods based on RGB cameras
are limited to RGB colors, we can choose to render modalities 
such as Lidar intensity, depth maps, and normals.
The final projection and voting step 
does not limit the number of votes, we therefore propose to train
multiple models on multiple modalities, and use them together in an ensemble.
We use the following modalities:
\begin{itemize}
    \item \textbf{Intensities}: The Lidar sensor intensity, normalized by scan, following \cite{caunes_3d_2024}.

    \item \textbf{Depths}: The point cloud depth maps with respect to the camera pose.
    \item \textbf{Normals}: The components of the point cloud surface normals, computed using 
    the normal estimation method in \cite{huang_neural_2023-1}.
\end{itemize}

For each modality, we generate a PC2D dataset, and train a separate model.
We then use ensembles of models to process each view of the scene at inference time.

\subsubsection{Projection and Fusion}
\label{method:projection}

\label{OCL}

To obtain final 3D segmentation masks from 2D models' outputs, 
we use a similar projection step as in \cite{caunes_3d_2024}, where 
2D logits are back-projected to 3D and accumulated as votes.
We propose adding occlusion to the projection step (OCL). 
Naive back-projection from 2D 
to 3D suffers from points being included in the projection even if they 
are occluded by objects.
We use depth maps generated from meshes, themselves generated using 
\cite{huang_neural_2023-1} for occlusion. Using meshes has the advantage 
of allowing less sparse depth maps which is more
 appropriate for dense point clouds.
Finally, we set a margin parameter $\delta=0.5m$ to include 
points behind the depth map to account for the size
 of the occluding object. 
 This final step yields the pseudo-labels for the target dataset.

%% file: figures/PC2D_generation.tex
\begin{figure}[th!]
\centering
\includegraphics[width=0.8\textwidth]{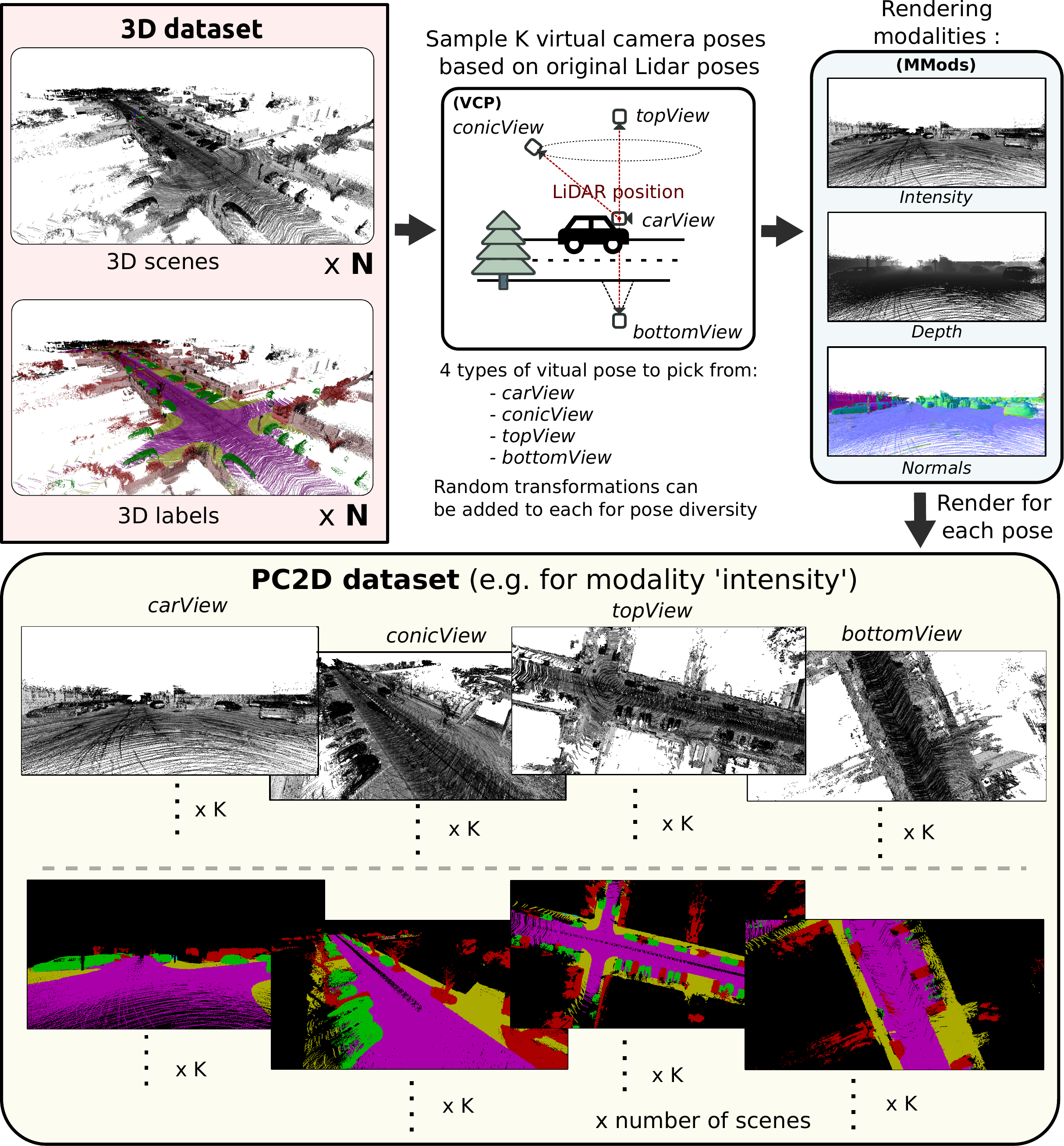}
\caption{\textbf{PointCloud2D (PC2D) dataset generation pipeline.} 
A 2D semantic segmentation dataset 
of rendered views of 3D point clouds is generated 
from a 3D semantic segmentation dataset.
To obtain diverse images, virtual camera poses of 4 categories 
are sampled around the 3D scenes.
The modality (colors) used for rendering can be chosen 
among sensor intensity, depth and projected normals.
For each scene and each camera pose category, a large number 
of camera poses are sampled, and the corresponding 2D images 
and segmentation masks are rendered.
}
\label{fig:PC2D_generation}
\vspace*{-1mm}
\end{figure}

%% file: sections/experiments.tex
\section{Experiments}

In this section we demonstrate the performance of our 
method for UDA on Lidar Semantic Segmentation tasks.

First, we show the performance of our method in 
both Real-to-Real and Simulation-to-Real settings.
For both settings, we compare the results with the state-of-the-art methods in UDA.
Then, we demonstrate how our method can be used to segment rare classes,
for which 3D annotations are not available, by only using 2D annotations 
for those classes and 3D annotations for other classes.
Finally, we conduct an ablation study to demonstrate the importance of
each component of our method.

\subsection{Datasets and Metrics}

\textbf{Real-to-Real.}
SemanticKITTI \cite{behley_semantickitti_2019} (SK) and nuScenes \cite{caesar_nuscenes_2020} (NS) are 
two well-known large-scale datasets for 
AD, both providing and annotating Lidar 3D point clouds. The procedure
generally adopted in the domain 
adaptation literature 
\cite{kim_single_2023,kim_rethinking_2024,gebrehiwot_t-uda_2023,shaban_lidar-uda_2023} 

is to train on the official training splits for both datasets, e.g. 700 scenes for nuScenes and
sequences 00 to 07, 09 and 10 for SemanticKITTI, and evaluate on the official validation splits, e.g.
150 scenes for nuScenes and sequences 08 for SemanticKITTI. We follow this procedure for comparing
with the state of the art. Of the 700 nuScenes scenes, we use the 70 last scenes for validation. 
Similarly, we use sequence 10 of SemanticKITTI for validation.
For the rare classes experiment \ref{exp:rare_classes} and the ablation study \ref{exp:ablation}, we
use a subset of the first 1800 scans from sequence 08 of the SK dataset for evaluation, 
to reduce the computational cost.

\textbf{Simulation-to-Real.}
SynLidar \cite{xiao_transfer_2021} (SL) is a synthetic AD dataset captured with Unreal Engine 4. 
The sub-sampled version that is used in most of the literature 
\cite{xiao_transfer_2021,saltori_compositional_2023,michele_saluda_2023,zhao_unimix_2024} 
is composed of 13 sequences, containing 19,865 scans
with annotations for 32 classes. No official validation split is provided, 
therefore we follow \cite{kim_single_2023} and use sequences 05 and 10 for validation.
We do not evaluate on SynLidar, as it is only used as a source dataset.

\textbf{Class mappings.}
We evaluate two Real-to-Real settings, \DAsetting{\ns}{\sk} and \DAsetting{\sk}{\ns},
and one Sim-to-Real setting, \DAsetting{\synth}{\sk}. To compare with the state-of-the-art,
we follow \cite{shaban_lidar-uda_2023,michele_saluda_2023}
and use 10 classes common to NS and SK.
For the \DAsetting{\synth}{\sk} scenario, we use the official SK19 class mapping as in 
\cite{saltori_compositional_2023,michele_saluda_2023}. To denote the different versions of
the SemanticKITTI dataset, we use the following notations: \skns, \sksyn.
The exact class mappings used can be found on github\footnote{https://github.com/andrewcaunes/ia4markings}.

\textbf{Metrics.}
We measure the 3D semantic segmentation performance using the 
Intersection over Union (IoU) metric for all the experiments.
We report both the per-class IoU and the mean IoU over all classes (mIoU).
We compute all the metrics using code provided by \cite{mmdetection3d_contributors_openmmlabs_2020},
to ensure a fair comparison.
For all UDA experiments, the evaluation is done with the 3D model 
after fine-tuning, to ensure similar conditions to the state-of-the-art methods,
i.e. real-time inference on single scans. 
For the rare classes
and ablation experiments, the evaluation is directly done on the pseudo-labels 
generated by the pipeline.

\subsection{Implementation details}
\label{exp:implementation}
The most important hyper-parameters are presented 
in this section, but more details can be found on 
github\footnotemark[\value{footnote}].

To begin with, we prepare the aligned scenes from SemanticKITTI and nuScenes. 
We use the already existing scenes 
from NS and accumulate scans 60 by 60 to create SK scenes. The intensities are 
normalized by scan, following \cite{caunes_3d_2024}.

NuScenes' annotations are only available at 2Hz, while the scans 
are captured at 20Hz. To prepare for 2D segmentation label generation for PC2D datasets, 
we propagate the annotations to all points with a simple KNN
method with $K=5$ to assign labels to each point in the dense scene.

The PC2D datasets are generated with 400,000 samples for each modality and source dataset.
The views are uniformly sampled over the training scenes.
For inference, we use 250 images per camera pose type, 
thus 1,000 images per scene, per modality. This amounts to 4,000 images total per scene.
All views are rendered at 1024$\times$512 resolution with a standard field of view.
The inference time for each scene takes \textasciitilde{} 2 minutes on 
a NVIDIA RTX 3070 GPU. This is however only done once for pseudo-label generation,
and there is no overhead introduced to the final fine-tuned model in both memory and time.

For the back-projection step, we crop the projection outside a depth range of [1m, 30m],
and when occlusion is used, we set the occlusion margin to $\delta=0.5m$ behind the depth 
encountered.

For the 2D segmentation models, we use the
Mask2Former \cite{cheng_masked-attention_2022} model 
with Swin backbone
\cite{liu_swin_2021}, from the framework of 
\cite{mmsegmentation_contributors_openmmlab_2020}. 
We pretrain the model on the 
PC2D dataset for 70,000 iterations 
on 1024$\times$512 resolution images with a batch size of 8. 
We use a fading learning rate 
of $1e^{-4}$.

For a fair comparison in UDA, we follow 
\cite{gebrehiwot_t-uda_2023,shaban_lidar-uda_2023,michele_saluda_2023} 
and use as 3D model a MinkowskiNet 32 from \cite{choy_4d_2019-2} 
as the 3D segmentation model with the framework of
\cite{mmdetection3d_contributors_openmmlabs_2020}.
We train it with the default settings from 
\cite{mmdetection3d_contributors_openmmlabs_2020} for $3\times 12$ epochs
on the source dataset, followed by $10,000$ iterations with batch size 4 
on the target pseudo-annotated dataset. The only augmentations used are 
basic random 3D flip, rotation, translation and scaling.

\subsection{Baselines}

To demonstrate the capabilities of our method, we compare it to the
state-of-the-art methods in UDA for 3D semantic segmentation. We use the results
reported by the authors for each method, while ensuring a fair comparison by using
the same dataset splits, class mappings, metrics and model architectures.
For Real-to-Real, in the \DAsetting{\ns}{\skns} setting, we find the 
best performing methods to date are 
Lidar-UDA \cite{shaban_lidar-uda_2023}, 
T-UDA \cite{gebrehiwot_t-uda_2023}, CoSMix \cite{saltori_compositional_2023} and 
SALUDA \cite{michele_saluda_2023}.
In the \DAsetting{\skns}{\ns} setting, we compare with Lidar-UDA \cite{shaban_lidar-uda_2023} 
and T-UDA \cite{gebrehiwot_t-uda_2023},
as CoSMix \cite{saltori_compositional_2023} and SALUDA \cite{michele_saluda_2023} do not provide results for this setting.
Note that T-UDA \cite{gebrehiwot_t-uda_2023} also reports values for the \textit{manmade} class,
not reported by the other methods. We therefore do 
not include it and re-compute the mIoU 
accordingly.
For Sim-to-Real, in the \DAsetting{\synth}{\sksyn} setting, we compare with 
SALUDA \cite{michele_saluda_2023}, CoSMix \cite{saltori_compositional_2023} and UniMix \cite{zhao_unimix_2024}.

\input{tables/table_NS2SK.tex}
\input{tables/table_SK2NS.tex}
\subsection{Unsupervised Domain Adaptation Results}
\label{exp:UDA}
\subsubsection{Real-to-Real setting}
Results for the \DAsetting{\ns}{\skns} setting can be found in Table
\ref{tab:uda_ns_sk}.
This UDA setting is particularly challenging as the source (\ns) uses a 32-beam Lidar,
less dense than the 64-beam Lidar used in the target (\sk).
\method{} is able to outperform the state-of-the-art by $+3.2$ mIoU points. 
Interestingly, our method reaches state-of-the-art performance even 
for dynamic classes such as \textit{car}, \textit{motorcycle} and \textit{truck}. This demonstrates that
the model learns to match `\textit{shadow effect}'-like features
with the corresponding classes.

Results for the 
\DAsetting{\skns}{\ns} setting can be found in Table
\ref{tab:uda_sk_ns}.
In this setting, \method{} is even more dominant, with a $+6.1$ mIoU advantage
over state-of-the-art.
The method results in particularly strong performance on static classes 
such as \textit{driveable surface},
\textit{sidewalk}, \textit{terrain} and \textit{vegetation}, which can be 
explained by the higher difficulty to segment dynamic classes displaying 
some `\textit{shadow effect}'.
For both tasks, we notice relatively lower performance on small classes, 
such as \textit{bicycle} and \textit{pedestrian}, which might be due to
the projection errors caused by poor sensor calibration being more important
for small objects.

\input{tables/table_SL2SK.tex}
\subsubsection{Simulation-to-Real setting}

Table \ref{tab:uda_sl_sk} reports the results for the
\DAsetting{\synth}{\sksyn} setting.
Overall, our method achieves lower mIoU compared to prior works such as 
CoSMix \cite{saltori_compositional_2023}, SALUDA \cite{michele_saluda_2023}, 
 and UniMix \cite{zhao_unimix_2024}. However, it consistently performs strongly on 
several static classes, notably, \textit{road} (80.2), \textit{sidewalk} (51.6), 
and \textit{building} (78.8), and some dynamic classes such as \textit{car} (83.5),
 where it reaches state-of-the-art performance. 
This suggests that the projection-based pseudo-labeling is 
particularly effective on large, well-structured classes but less robust 
on fine-grained or rare classes such as \textit{bicycle}, \textit{motorcycle}, 
and \textit{motorcyclist}. 
While the global mIoU is lower (28.2), the results highlight that our approach 
can still surpass prior methods on key static categories, underlining its 
strength for Sim-to-Real in some applications such as road infrastructure 
analysis.

\subsection{Ablation Study}
\label{exp:ablation}
We conduct a complete ablation study to verify the importance of each component of our method.
Results can be seen in Table \ref{tab:ablation}.
The first stage (a) corresponds to the base multi-view projection method that we 
followed from \cite{caunes_3d_2024}.
We add the PC2D dataset training in stage (b), 
the occlusion handling in stage (c),
the Virtual Camera Placement in stage (d), 
and the full \method{} with the Multi-Modalities module
is evaluated in stage (e).
We evaluate the models in both \DAsetting{\skns}{\skns} (in-domain) and \DAsetting{\ns}{\skns} (UDA).

\input{tables/table_ablation.tex}

We see that the addition of the PC2D dataset leads to the most significant improvement,
followed by the occlusion handling and the multi-modalities module.
Interestingly, the addition of the VCP module leads to an improvement only for
the \DAsetting{\skns}{\skns} setting, and not for UDA. This might be because
using more diverse camera poses increases the risk of encountering a large 
domain shift within one camera pose type, e.g. the difference in Lidar point density
might yield a larger difference in BEV type views than in \textit{road user} type views,
since the difference in point density becomes significant when seen
from afar.

Overall, all components proposed in our method seems valuable for either supervised or UDA applications.

\subsection{Application to rare classes}

A common real world scenario in 3D semantic segmentation is to have 
access to 3D annotations for some classes in a source domain, but not the 
desired target classes.
However, 2D annotations are more generally available as they are less costly. 
For such a situation when 2D annotations of the target classes are available, 
\cite{caunes_3d_2024} proposes a multi-view 
projection method with 2D models trained on real camera images,
which then perform out-of-domain inference on virtual rendered 
views.
We show that our method can help to significantly 
improve the performance of this approach
by using the 3D annotations and our PC2D dataset 
generation pipeline to pretrain the 2D model
in rendered view domain. The model is then fine-tuned on 
real camera images with annotations for the target classes.
The pretraining on 3D annotations allows to significantly
bridge the domain gap that exists at inference time by 
first exposing the model to rendered images.

\subsubsection{Experimental setup}

For this experiment, 3 datasets are used:
\begin{itemize}
    \item A \textit{pretraining} 3D source dataset, with 3D annotations for any classes 
    (but not the target classes). We use nuScenes (NS)~\cite{caesar_nuscenes_2020}.
    \item A \textit{rare classes} 2D dataset, with 2D annotations for the desired target classes. We 
    use Mapillary Vistas (MV)~\cite{neuhold_mapillary_2017}, using only the \textit{traffic sign} class
    as the target class.
    \item A \textit{target} 3D dataset. We use SemanticKITTI (SK)~\cite{behley_semantickitti_2019}, 
    as it conveniently provides 3D annotations
    for our target class, \textit{traffic sign}, which will allow to perform a quantitative evaluation.
\end{itemize}
We compare our \method{} with the baseline from \cite{caunes_3d_2024}, 
both trained on the \textit{rare classes} dataset.
Differently from \cite{caunes_3d_2024} and from our base method in Sec.~\ref{method}, 
we use the RGB colorized point clouds using camera images 
for SK and NS as the rendered views' modality
instead of Lidar intensities or other modalities.
This reduces the domain gap with 
the real camera images of the \textit{rare classes} dataset,
but requires that camera data be available for the \textit{pretraining} dataset.
We use Mask2Former models for both methods with the same parameters as \ref{exp:implementation}.
For \method{}, we pretrain the 2D model on a PC2D dataset, generated from the \textit{pretraining} dataset,
 as in \ref{exp:implementation},
and then fine-tune the model on the \textit{rare classes} dataset for 10,000 
iterations, while freezing the first block of the Swin backbone.

\input{tables/table_rare_classes.tex}
\subsubsection{Results}
\label{exp:rare_classes}
Table \ref{tab:rare_classes} reports the results as computed on
SK for the \textit{traffic sign} class. We see a significant improvement over the base method,
with a $+6.4$ mIoU improvement, corresponding to relative increase of $+53\%$. 
This shows that our method can be used to segment rare classes for which
3D annotations are not available, when some 2D annotations are available.

%% file: tables/table_NS2SK.tex
\begin{table}[t!]
    \small
    \centering
    
    \newcommand*\rotext{\multicolumn{1}{R{45}{2em}}}
    \setlength{\tabcolsep}{3.5pt}
    
    \makebox[\linewidth][c]{
    \begin{tabular}{l|cccccccccc|c}
    \toprule
    
    \DAsetting{\ns}{\skns}
    
    (\%\,IoU)
     & \rotext{Car} &	\rotext{Bicycle}	    & \rotext{Motorcycle} &\rotext{Truck} &	\rotext{Other vehicle}	   & \rotext{Pedestrian}	 & \rotext{Driveable surf.} &	\rotext{Sidewalk} &	\rotext{Terrain} &\rotext{Vegetation} & \rotext{\%\,mIoU}\\ \midrule
    \rowcolor{black!10}
    Source-only  & 74.1  & 0.0 &10.2 & 1.1 & 1.0 & 4.4 & 63.4 & 31.8 & 36.0 & 25.6 & 24.8  \\ 
     
     T-UDA  \cite{gebrehiwot_t-uda_2023}      & \best{93.0}    & 0.0           & 11.4          & 3.4           & \best{47.0}    & 15.7           & \best{83.3}   & \second{54.4}     & \second{67.9} &  83.9         & 46.0         \\ 
     Lidar-UDA \cite{shaban_lidar-uda_2023}   & 86.2          & 0.0          & 13.9         & 9.3          & 3.1            & 16.5          & 65.7         & 6.1            & 54.1         &  85.7        & 34.1        \\  
     CoSMix \cite{saltori_compositional_2023} & 77.1     	     & \second{10.4} & 20.0	         & 15.2	       & 6.6            & \best{51.0}    & 52.1          & 31.8            & 34.5          &  84.8         & 38.3         \\  
     SALUDA \cite{michele_saluda_2023}        & 89.8           & \best{13.2}   & \second{26.2} & \second{15.3} & 7.0	        & \second{37.6}           & 79.0 & 50.4            & 55.0          & \second{88.3} & \second{46.2} \\

     \midrule
    Ours & \second{90.1} & 0.0          & \best{28.4}  & \best{17.7}  & \second{18.4}	& 37.1 & \second{80.9} & \best{61.0}  & \best{70.2}  & \best{90.1}  & \best{49.4}  \\

    \bottomrule

    \end{tabular}}
    \vspace*{1mm}
    \caption{\textbf{UDA, \DAsetting{\ns}{\skns}. IoU per class and mIoU.} Color:\colorbox{red!20}{\color{black}Best}, \colorbox{red!10}{\color{black}Second}. 
    }
    \label{tab:uda_ns_sk}
    \end{table}

%% file: tables/table_SK2NS.tex
\begin{table}[t!]
    \small
    \centering
    
    \newcommand*\rotext{\multicolumn{1}{R{45}{1em}}}

    \setlength{\tabcolsep}{3.5pt}
    
    \makebox[\linewidth][c]{
    \begin{tabular}{l|cccccccccc|c}
        \toprule
    
     \DAsetting{\skns}{\ns}

     (\%\,IoU)
     & \rotext{Car} &	\rotext{Bicycle}	    & \rotext{Motorcycle} &\rotext{Truck} &	\rotext{Other vehicle}	   & \rotext{Pedestrian}	 & \rotext{Driveable surf.} &	\rotext{Sidewalk} &	\rotext{Terrain} &\rotext{Vegetation} & \rotext{\%\,mIoU}\\ \midrule
    \rowcolor{black!10}
    
    Source-only & 1.8 & 0.0 & 0.0 & 0.0 & 0.0 & 0.5 & 6.7 & 0.1 & 0.1 & 0.3 & 0.9 \\  
     
     T-UDA   \cite{gebrehiwot_t-uda_2023}      & \best{74.2}  & \second{0.5} & \best{40.3} & \second{21.8} & \second{0.2} & 0.4 & \second{87.8} & \second{45.8} & 46.1 & 70.3 & 38.7 \\ 
     Lidar-UDA  \cite{shaban_lidar-uda_2023}   & \second{73.5} & \best{0.9}  & 15.9 & 0.9    & \best{25.7} & \best{40.8}& 87.4 & 42.3 & \second{47.9} & \second{83.2} & \second{41.8} \\

     \midrule
     
     Ours               & 73.3  & \second{0.5} & \second{32.0} & \best{50.7} & 0.0 & \second{32.1} & \best{90.3} & \best{55.4} & \best{58.2} & \best{87.0} & \best{47.9} \\
    
    \bottomrule
    \end{tabular}
    }
    \caption{\textbf{UDA, \DAsetting{\skns}{\ns}. IoU per class and mIoU.} Color:\colorbox{red!20}{\color{black}Best}, \colorbox{red!10}{\color{black}Second}.}
    \label{tab:uda_sk_ns}
\end{table}

%% file: tables/table_SL2SK.tex
\begin{table}[!t]
    
    \tiny

    \centering
    
    \newcommand*\rotext{\multicolumn{1}{R{90}{-0.3em}}}
    \setlength{\tabcolsep}{1pt}
    
    \makebox[\linewidth][c]{
    \begin{tabular}{lc|cccccccccccccccccccc}
    \toprule
     \rlap{\raisebox{12mm}{\synthtosk}}
     \rlap{\raisebox{6mm}{~~(\%\,IoU)}}
     && \rotext{Car} &	\rotext{Bicycle} & \rotext{Motorcycle} & \rotext{Truck} &	\rotext{Other vehicle} &	\rotext{Pedestrian} &	\rotext{Bicyclist} &	\rotext{Motorcyclist} & \rotext{Road} & \rotext{Parking} & \rotext{Sidewalk} &	\rotext{Other ground} &	\rotext{Building} &	\rotext{Fence} & \rotext{Vegetation} &	\rotext{Trunk} & \rotext{Terrain} &	\rotext{Pole} &\rotext{Traffic sign} & \rotext{\%\,mIoU}\\ \midrule

    \rowcolor{black!10}

    Source-only && 50.8 & 7.6 & 11.8 & 1.4 & 3.2 & 15.8 & 34.8 & 3.9 & 30.0 & 4.5 & 34.5 & 0 & 27.0 & 15.8 & 62.9 & 24.4 & 43.0 & 18.9 & 6.4 & 20.9 \\ \midrule

    CoSMix & \cite{saltori_compositional_2023} & 75.1 & \second{6.8} & \second{29.4} & \second{27.1} & \best{11.1} & 22.1 & 25.0 & \best{24.7} & \second{79.3} & \second{14.9} & \second{46.7} & \second{0.1} & 53.4 & 13 & 67.7 & 31.4 & 32.1 & 37.9 & 13.4 & \second{32.2} \\

    SALUDA & \cite{michele_saluda_2023} & 67.0 &	\best{7.7}	&	14.4&	1.3	&	5.2		& \best{24.1}	&	 \best{52.6}	 &	2.7	 &	52.5	 &	10.5	 &	44.1	 &	\best{0.4}		 & 51.8	 &	13.6 &		\second{69.7}	 &	\best{40.5}	 &	\best{56.5}	 &	\best{45.0}	&	\second{14.3} & 30.2 \\
    Unimix & \cite{zhao_unimix_2024}   & \second{80.3} & 6.1 & \best{32.5} & \best{29.2} & \second{10.6} & \second{23.7} & \second{30} & \second{24.5} & 62.2 & \best{15.9} & 46.5 & \second{0.1} & \second{60.9} & \second{15.8} & \best{70.7} & \second{34.9} & \second{41.2} & 38.6 & \best{17.1} & \best{33.7} \\
    \midrule

     Ours &                                          & \best{83.5} & 0.9 & 6	 & 10.4 & 3.0 & 19.8 & 13.5 & 0.2 & \best{80.2} & 0.0 & \best{51.6} & 0 & \best{78.8} & \best{16.7} & 51.9 & 32.0 & 36.8 & \second{40.6} & 10.1 & 28.2  \\ 
     \midrule 
    \end{tabular}}
    \caption{\textbf{UDA, \synthtosk. IoU per class and mIoU.} Color:\colorbox{red!20}{\color{black}Best}, \colorbox{red!10}{\color{black}Second}.}
    \label{tab:uda_sl_sk}
\end{table}

%% file: tables/table_ablation.tex
\begin{table}
    \small
    \setlength{\tabcolsep}{2pt}
    \centering
    \begin{tabular}{c|cccc|ccccc}
    \toprule
    \multicolumn{1}{l}{}&PC2D&OCL&VCP&MMods&\skns& $\nearrow_{mIoU}$ &\ns & $\nearrow_{mIoU}$ \\
    
    \midrule
    \rowcolor{blue!3}
    (a)& & &  & & 15 & \textcolor{mygreen}{-} & 15 & \textcolor{mygreen}{-} \\
    (b)&\cmark& &  & & 31.1 & \textcolor{mygreen}{+16.1} & 28.9 & \textcolor{mygreen}{+13.9}  \\
    (c)&\cmark&\cmark&  & & 42.0 & \textcolor{mygreen}{+10.9} & 34.4  & \textcolor{mygreen}{+5.5} \\
    (d)&\cmark&\cmark&\cmark & & 43.7 & \textcolor{mygreen}{+1.7} & 34.3  & \textcolor{mygreen}{-0.1} \\
    \rowcolor{blue!10}
    (e)&\cmark&\cmark&\cmark&\cmark & 49.3 & \textcolor{mygreen}{+5.6} & 40.9 & \textcolor{mygreen}{+6.6}  \\
    
    \bottomrule
    \end{tabular}
    
    \vspace*{2mm}
    \caption{\textbf{Ablation study.} (f) is full \method{}. (a) is \cite{caunes_3d_2024}. 
    Models are all evaluated on \skns. Datasets indicated for 
    the results columns are the training datasets.
    Results are in \% mIoU.}
    \label{tab:ablation}
    \end{table}

%% file: tables/table_rare_classes.tex
\begin{table}[t!]
    \small
    \setlength{\tabcolsep}{3pt}
    \centering
        \begin{tabular}{l@{}r|c|c}
            \toprule

             Method & &  traffic sign & $\nearrow_{mIoU}$ \\        
            \midrule
            Baseline & \cite{caunes_3d_2024} & 11.9 & \\ 
            \rowcolor{blue!5}
            \method{}~~~~~~~ & \llap{(ours)} & \textbf{18.3} & \textcolor{mygreen}{+6.4 (+53\%)} \\
            
            \bottomrule
        \end{tabular}
   
        \vspace*{2mm}
    \caption{\textbf{Application to rare classes.} IoU on SemanticKITTI for the `traffic sign' class. }
\label{tab:rare_classes}
\end{table}

%% file: sections/conclusion.tex
\section{Conclusion}
We have presented a new UDA method based on a novel 
multi-view projection framework for 3D semantic segmentation. 
After generating 
large-scale synthetic 2D datasets (PC2D) from aligned Lidar scenes
in the source domain to train an ensemble of 2D segmentation models, 
our method leverages occlusion-aware back-projection and voting to 
produce high-quality 3D pseudo-labels for the target domain.

Extensive experiments demonstrate that our approach achieves 
state-of-the-art performance in Real-to-Real UDA and
useful performance for large classes in Sim-to-Real UDA.
We also demonstrated a potential application of our method
to segment rare classes with no 3D annotations but 2D annotations
available.

While our approach thoroughly explored problem $1)$ mentionned in Section \ref{related:multi}
of what views to use for inference, and which models to infer with, 
future research could be led to explore problem $2)$
of how to optimally use the numerous outputs 
from the ensemble of models on all views 
to assign final 3D labels.
In particular, our logits accumulation and voting scheme does not use 
important information such as spatial proximity of the points or class size statistics.
Information like this could be used to train a model to output 
the final 3D segmentation mask with more insight than a 
simple voting scheme.

We hope that our work will inspire further research into this direction.